\documentclass{article}
\usepackage{ijcai20}

\usepackage{times}
\usepackage{soul}
\usepackage{url}
\usepackage{multirow}
\usepackage[utf8]{inputenc}
\usepackage[small]{caption}
\usepackage{graphicx}
\usepackage{hyperref} 
\usepackage{amsmath}
\usepackage{amsthm}
\usepackage{booktabs}
\usepackage{algorithm}
\usepackage{algorithmic}
\title{Exploiting Latent Codes: Interactive Fashion Product Generation, Similar Image Retrieval, and Cross-Category Recommendation using Variational Autoencoders}

\author{
    James-Andrew R. Sarmiento
    \affiliations
    Department of Computer Science, University of the Philippines, Diliman
    \emails
    jrsarmiento1@up.edu.ph
}

\begin{document}

\maketitle

\begin{abstract}
The rise of deep learning applications in the fashion industry has fueled advances in curating large-scale datasets to build applications for product design, image retrieval, and recommender systems. In this paper, the author proposes using Variational Autoencoder (VAE) to build an interactive fashion product application framework that allows the users to generate products with attributes according to their liking, retrieve similar styles for the same product category, and receive content-based recommendations from other categories. Fashion product images dataset containing eyewear, footwear, and bags are appropriate to illustrate that this pipeline is applicable in the booming industry of e-commerce enabling direct user interaction in specifying desired products paired with new methods for data matching, and recommendation systems by using VAE and exploiting its generated latent codes.
\end{abstract}

\section{Introduction}
The rise of machine learning and deep learning applications in the fashion industry has fueled more progress in curating large-scale datasets such as \cite{fashionai,deepfashion}. These advances in the availability of huge datasets have been the starting point of new methods in using deep learning for integration in e-commerce for the fashion industry to adapt to the ever-changing shoppers' preferences and to increase profits by being more collaborative with the customer in terms of learning from their buying patterns. Recommendation engines have been one of the most heavily studied topics in the areas of computer science, specifically in machine learning and artificial intelligence because of its tangible use in the daily lives of users aiming to personalize user experience by modelling their preferences. In addition, it also allows users to discover new content within the platform. One such example is recommending a new movie to watch in an online streaming platform. 

Given all of the possible applications of recommender systems such as social media, streaming platforms, e-commerce websites, dating applications, and the rest, most of the research has been directed towards a specific domain. In this paper, the author focuses on the applicability of recommender systems specifically for fashion product recommendations. In the realm of online shopping, users would want to see products similar to their tastes, such as different pant designs or sneakers. At the same time, these users would also want to be able to mix and match different fashion products not far from their own tastes. Hence, an interactive fashion product application is developed that allows the users to generate products with attributes according to their liking, retrieve similar styles for the same product category and also allow users to perform cross-category recommendations which might be of interest for users who would want to buy a set of fashion products according to their designed look.

The author proposes using Variational Autoencoders (VAE) where the author exploits its capability to regularise the latent space for generating new samples from the distribution and to generate a latent representation of data as features and introduces two methods in utilizing the latent variables for similar image retrieval, recommendation. This enables a base framework that can be used for the e-commerce industry that implements collaborative design for product styles, product clustering, image retrieval and recommender systems using only one architecture to solve multiple problems.

\section{Related Work}
\subsection{Variational Autoencoders}
Variational Autoencoders (VAEs) are similar to the structure of autoencoders but with a different approach for latent representation learning. An autoencoder (AE) is a type of Artificial Neural Network (ANN) where the expected output is exactly the input by finding an efficient latent representation of the data or encodings using a hidden or latent layer called \textit{"bottleneck"} layer (often called latent codes in this paper). Both AE and VAE have similar structures made of an encoder that maps the input $x$ into a latent space $z$, the bottleneck layer, and a decoder that takes $z$ to reconstruct the input data into $x'$, in which the target is to minimize the reconstruction loss or the difference between $x$ and $x'$ \cite{designingvae}. The main challenge for autoencoders in general is to accurately learn a representation that embeds the most important features of the data and encode them to generalise beyond the training set by being able to gain control over the latent space to produce images with similar characteristics.

Instead of using the output of the encoder (i.e. the bottleneck), VAEs allow generative modelling by causing the latent variables to be normally distributed by forming a new layer in the architecture consisting of parameters mean $\mu$ and standard deviation $\sigma$ of a normal distribution $\mathcal{N}$($\mu,\sigma^2$) \cite{towardsvae}. During optimization and training, the target reference is is a standard normal distribution $\mathcal{N}$(0,1) where we force our normal distribution to diverge as close as possible to our target reference using Kullback-Leibler (KL divergence) \cite{tutorialderive}. VAE now calculates loss function consisting two terms - KL-loss, which is dependent on the parameters of the distribution and can be computed easily, and reconstruction loss which is based on random and discrete outcome of $z$ that means its gradients are not computable preventing backpropagation over parameters $\mu$ and $\sigma$.

\begin{equation}\label{epsilonsampling}
    z = \mu + \sigma^2 \odot \epsilon, \epsilon \sim Normal(0,1)
\end{equation}

\begin{figure}[h!]
\centering
\includegraphics[width=0.4\textwidth]{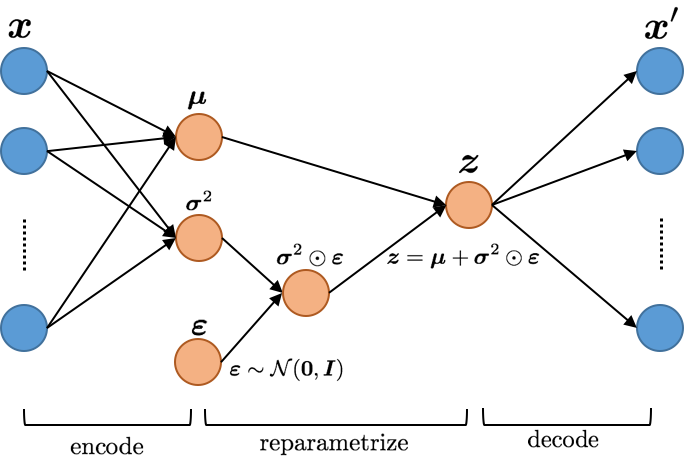}
\caption{Variational Autoencoder Architecture adopted from \protect\cite{ReNom}. $\mu$, $\sigma$, $\epsilon$, and $z$ are all n-dimensional vector. $x$ is the input image vector and $x'$ is the image reconstruction}
\label{fig:vaeimg}
\end{figure}

To solve this, \cite{autoencoding} proposed a method called \textit{reparameterization trick} which tries to transform and approximate a normal distribution $\mathcal{N}$($\mu,\sigma^2$) into $\mathcal{N}$(0,1), allowing calculation of the gradients over $\mu$ and $\sigma$ by introducing a random component $\epsilon$ sampled from $\mathcal{N}(0,1)$ as seen in Equation \ref{epsilonsampling} which eliminates randomness since the last remaining random entity in the equation is $\epsilon$. This enables the training of VAE to achieve both the generative and feature extraction capability applicable on this pipeline. Techniques and methods in this paper focus on how to take advantage of this inherent structure in VAE. The overall structure of VAE is seen in Figure \ref{fig:vaeimg}.

\subsection{Fashion Product Generation}
The fashion industry has been quick to adapt in designing new products using deep learning. Two popular deep learning models for generative modelling are Generative Adversarial Networks (GANs) and Variational Autoencoders (VAEs). \cite{shoegan} experimented with a traditional GAN and DiscoGAN trained on athletic shoes using UT-Zap50K dataset and tested the generated images using a classifier based on GoogleNet and achieving 88\% validation accuracy. \cite{fashiongen} created a baseline results on high-resolution image generation conditioned on the given text description using their own created dataset of 293,008 high definition fashion images paired with item decriptions. 
Variational Autoencoders are also popular in generating new product designs. \cite{vaegeneration} used VAE on popular fashion image styles trained on DeepFashion dataset and was able to generate new products based on 30 discovered styles using Nonnegative matrix factorization (NMF). 
Current generative modelling puts emphasis on synthesizing new product designs that are realistic but enabling the end-user to be the one interacting with the model to create desired designs are still missing.

\subsection{Image Retrieval}
Deep learning has been used in content-based image retrieval and matching for quite a long time such as for medical images \cite{medicalimage} which uses deep Convolutional Neural Network (CNN) achieving 99.97\% classification accuracy for retrieval tasks and for clothing products that uses CNN to match a real-world example of garment into same item in an online shop.

Traditional autoencoders and variational autoencoders are often used for image reconstruction or generation, but they can be used for image retrieval because of its capability to extract features. \cite{autoencoderretrieval} used a very deep autoencoder and semantic hashing for image retrieval by mapping images to short binary codes for extreme fast retrieval, while \cite{designingvae} implemented a simple VAE with only 2 layers depth in both encoder and decoder and yielded good results using approximation of the latent space to $z$ = $\mu_z$ evaluated by mAP. 

\subsection{Recommendation Systems}
Content-based Recommendation has been dominated by deep learning where there is emphasis on finding similarity on two images using deep neural networks. \cite{cnnrecom} used two convolutional neural network where the first neural network is used for classification and the second neural network to model similarity score between pair of images using Jaccard similarity. This method was able to achieve 0.5 accuracy on both classification and similarity score. Another work by \cite{cnndeepreco} utilized a combination of CNN and a modified k-NN algorithm used for ranking in feature space. GANs can also be used to recommend products. A work by \cite{cgan} implemented $c^+$GAN which tweaks traditional c-GAN where it recommends a generated pair of pants that might go well with the given shirt. 

A work by \cite{farmoutfit} proposed a neural co-supervision learning framework which they called FAshion Recommendation Machine (FARM) which is a neural network architecture similar to variational autoencoders that improves visual understanding and enhances visual matching by incorporating a layer-to-layer matching mechanism for combining aesthethic information. This was able to outperform state-of-the-art models for outfit recommendation. Works on cross-domain recommendation typically revolve around recommending one category.

Researches in fashion product generation, content-based similar image retrieval and recommendation have utilized deep learning to an extent but each study only focus on solving one problem using one deep learning architectures. In this work, the author experimented on using VAE by exploiting its ability to produce a latent representation of the image for generative modelling and feature extraction and introduced new ways to utilize these latent codes as features to solve all the problems aforementioned that are related in nature and can be integrated in a pipeline for fashion industry.

\section{Proposed Approach}
Many machine learning technologies have been applied for generating new designs for products, matching images on databases, and recommendation systems. These architectures are utilized on specific single use-cases to solve one problem.

\begin{figure}[h!]
\centering
\includegraphics[width=0.5\textwidth]{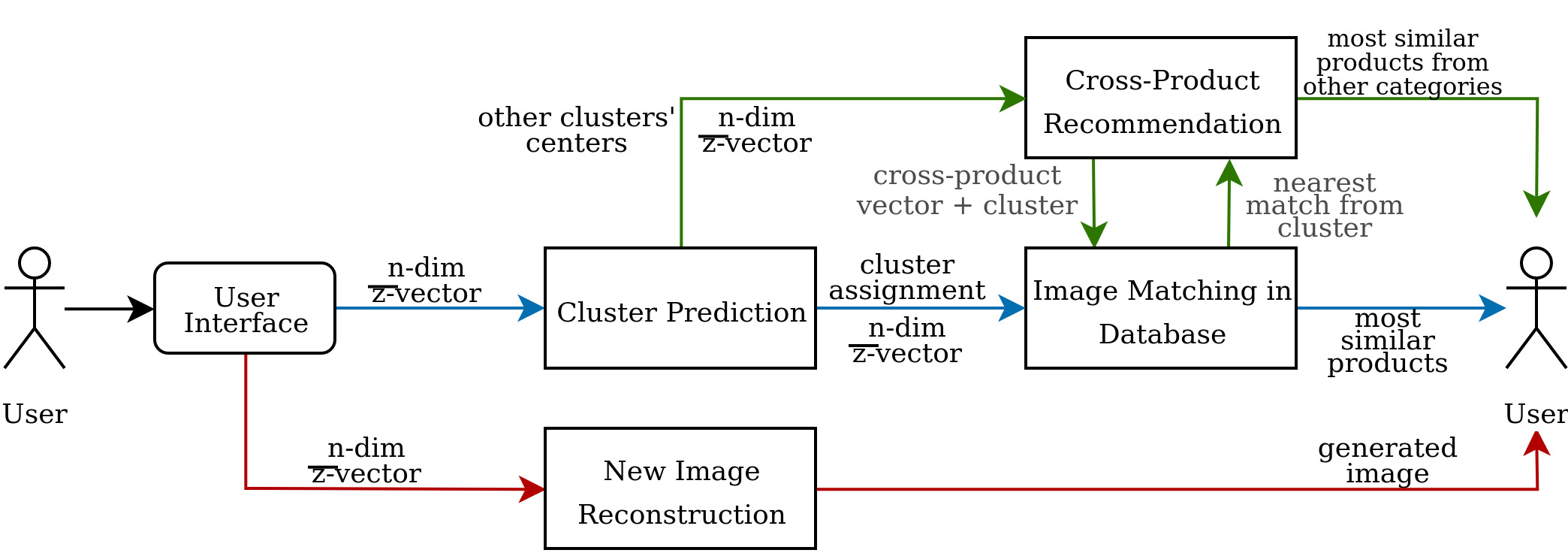}
\caption{Simplified Diagram for the pipeline of generating new images, finding most similar products, and recommending products from other categories.}
\label{fig:diagram}
\end{figure}

The simplified diagram is shown in Figure \ref{fig:diagram} where it illustrates how the user-generated $\bar{z}$-vector is used in the entire pipeline. This section discusses how user interaction with the application and controlling the latent code generates new product images and how this pipeline capitalizes on the use of the given latent code variables as features in clustering, similar image retrieval, and recommendation. 

\subsection{Dataset}
The dataset is obtained from the combination of Fashion Product Images Dataset from Kaggle and scraping images from Zappos website. The dataset is intentionally limited to footwear (8919 entries), eyewear (3378 entries) and bags (9462 entries) which totals to 21,759 data samples. Topwears and bottomwears are not considered given that majority of topwear and bottomwear includes human faces and poses that could add entropy and more features for the variational autoencoder to represent. The size of the images are 180 x 240 pixels.

\subsection{Training}
The model is trained in Google Colab that uses K80 Tesla as GPU and includes 25GB of RAM for 200 epochs which took approximately four hours. The model architecture is composed of three Convolution layers and Dense layers for the encoder and three Deconvolution layers and Dense layers for the decoder and is built in Keras. The bottleneck layer is size 16.

\subsection{Interactive Synthesizing of New Fashion Products}
Here, the user is able to directly control the z-vector that is used to synthesize new images by using range sliders. By controlling the range slider, the user is able to manipulate certain attributes such as length, color, thickness, etc. of the fashion products according to their preference. As the range slider is moved, the latent code is changed and is given to the decoder in the back-end of the application for reconstructing the image. For consistency, the user-generated vector from using the application sliders is denoted as $\bar{z}$-vector.

\subsection{Similar Image Retrieval Using Latent Codes}
After the user interacts with the slider and synthesizes the image, its $\bar{z}$-vector is used to find similar images from the database. For this purpose, the author proposes two methods that tries to solve the problem in different manners. The first one takes advantage of the objective function being optimized by VAEs while the second one exploits the sampling process from reparameterization trick used to derive $z_x$ from $\mu_x$ and $\sigma_x$. 

In order to use these techniques for image retrieval, encodings of each images are generated using the trained encoder and stored in a database. The encoding for an image is composed of n-dimensional $\mu_x$-vector and $\sigma_x$-vector and an n-dimensional vector computed from a fixed epsilon sampling that will be discussed in the next sections.

\subsubsection{Log-Likelihood Maximization in Image Retrieval}
Distance metrics are used in calculating the similarity of two points. In this image retrieval technique, the objective is to find the closest images in the database by calculating the distance between the user-generated $\bar{z}$ vector and the $\mu_x$ and $\sigma_x$ encoding vectors generated earlier.

On training, the encoder is essentially parameterizing the variational posterior distribtion of $z$ given $x$, (i.e. Equation \ref{loglikelihood}. Think of the problem as if the $x$ are parameters  of the probability distribution while $z$ are some observations you made.  This implies that the nearest encoding  from the training data $x$ would  be the encoding with the highest likelihood evaluated by the equation taking the $x$ with the maximal value. The log-likelihood provides something similar to a negative euclidean distance between $\mu(x)$ and $z$, but scales and shifts by terms determined by the standard deviation.

\begin{equation}\label{loglikelihood}
\resizebox{.90\linewidth}{!}{$
    \displaystyle
    \log q(z|x) = \sum_{i=1}^{N_z} \log q(z_i|x) = \sum_{i=1}^{N_z} \left[ -\frac{(z_i - \mu_i(x))^2}{2\sigma_i(x)^2} - \log \left( \sqrt{2\pi}\sigma_i(x) \right) \right]
$}
\end{equation}

Intuitively, by maximizing the log-likelihood using Equation \ref{loglikelihood}, you are minimizing the euclidean distance between a given $z$, which in this case is from the user interaction with the application, and the encoding $\mu(x)$ of $x$ from the encoding of each images stored. 

\subsubsection{Fixed Epsilon Sampling Encoding Trick in Image Retrieval}
The problem that these methods are trying to solve is finding a way to calculate the difference between two image encodings given that (1) the encoder only gives $\mu_x$ and $\sigma_x$ vectors, (2) a random sampling is involved to get the $z$-vector of the image, and (3) that the user only provides us a $\bar{z}$-vector. 

The biggest problem is that in order to generate a $z$-vector encoding of each images, random sampling of $\epsilon$ from a Gaussian distribution is involved as seen in Equation \ref{epsilonsampling}. This random sampling implies different variations of $\epsilon$ used to calculate $z$. The previous method considers computing the distance between the user-generated $\bar{z}$-vector and the $\mu_x$ and $\sigma_x$ vector encodings of each images in the database, in which case, there is no $\epsilon$-sampling required.

The trick is to use a fixed, n-sized $\epsilon$ vector randomly sampled once from a Gaussian distribution for each encoding process of the images. In other words, instead of randomly sampling $\epsilon$ for every encoding process, the $\epsilon$ is held constant and $z$ is calculated using the same Equation \ref{epsilonsampling}. This method allows us to directly measure the distance between $\bar{z}$-vector and each encoded $z$-vector in the database using any traditional distance metric such as Euclidean distance, Manhattan Distance, etc.

\subsection{Clustering Images using its Latent Codes}
Image retrieval is computationally expensive, most especially if the algorithm has to compare the $\bar{z}$-vector to every entry in the database. In order to reduce the number of comparisons, images could be clustered using their generated latent codes. Creating clusters also allows us to find a method of recommending products from other clusters and categories that also uses the latent encodings. 

To cluster each products, K-Means Clustering can be used as it allows choosing the number of $k$ clusters to group the data and provides the value of the centers for each clusters. These centers will be used for cross-product recommendation. In K-Means clustering, given a set of n data points in a high-dimensional space and integer $k$, the objective is to determine $k$ points called centers that minimizes the mean squared distance from each point to its nearest designated center\cite{kmeans}. Since the dataset comes with a subcategory tag, $k$ would be equal to three.

The author experimented with Euclidean distance as metric for K-means and evaluated using $\mu_x$-vector in comparison to using the fixed epsilon sampled $z_x$-vectors. Also, before the image retrieval process, the image constructed from the user-generated $\bar{z}$-vector is assigned to a cluster by calculating the distance in comparison to the stored cluster centers. 

\subsection{Cross-Product Recommendation using Latent Codes}
Aside from providing the most similar images to the user, latent codes can also be employed in recommending products from other clusters or categories. This also takes full advantage of the nature of the latent space VAEs generate which enables smooth interpolation and smooth mix of features a decoder can understand.


Simple vector arithmetic in the latent space can be utilized to navigate through the smooth interpolation of the latent space. For example, to generate specific features such as generating glasses on faces, we find two samples, one with glasses and one without and obtain the difference of their encoded vectors from the encoded. We add the difference as "glasses" vector to the encodings of any other image that has no glasses.

\begin{equation}\label{crossproduct}
    recommendation_{ij} = \bar{z} + diff_{ij}
\end{equation}

Here, the clusters or categories are defined by the cluster centers. Take the difference of each center vectors (e.g. $center_i$ - $center_j$) as $diff_{ij}$ from one another where $i$ is the cluster assignment of $\bar{z}$-vector and $j$ is a cluster not equal to $i$. In order to recommend products from other clusters, add each difference vector $diff_{ij}$ to $\bar{z}$-vector as seen in Equation \ref{crossproduct}. 

This does not recommend a reconstructed image using the decoder with $recommendation_{ij}$. It does, however, recommend a retrieved image from the database with the nearest to the calculated vector from the $j$th cluster using the presented image retrieval techniques.

\section{Results and Discussion}
\subsection{Fashion Product Image Synthesis}
\begin{figure}[h!]
\centering
\includegraphics[width=0.4\textwidth]{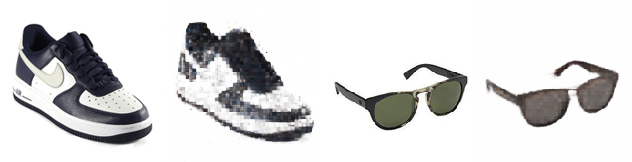}
\caption{Sample Image Reconstruction after training}
\label{fig:vaetraining}
\end{figure}

Figure \ref{fig:vaetraining} shows comparison of the original image and the reconstruction of the image using the trained model. VAEs are known to produce blurry images due to the small bottleneck layer and latent space, but the reconstruction is still very recognizable as the original. The reconstruction of the first image in Figure \ref{fig:vaetraining} caught the right colors and structure but missed out some intricate details while the reconstruction of the glasses had some washed out colors.

\begin{figure}[h!]
\centering
\includegraphics[width=0.35\textwidth]{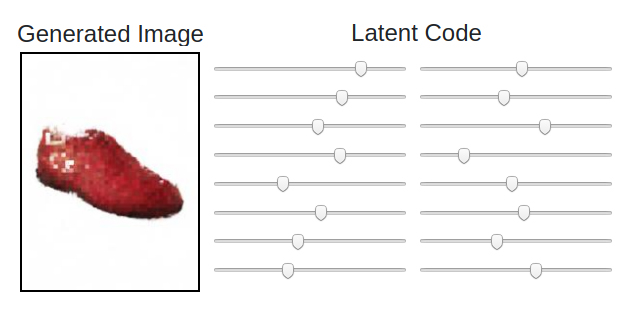}
\caption{Interactive Sliders for Controlling $\bar{z}$ latent vector}
\label{fig:sliders}
\end{figure}

Since the bottleneck layer is size sixteen, the user is provided sixteen range sliders for synthesizing images by manipulating the $\bar{z}$-vector. This is decoded by the decoder in the backend. The application provides the user an encoding from the database selected randomly as starting point. The user is given sixteen sliders to interactively control the $\bar{z}$ latent vector (Fig. \ref{fig:sliders}).

\begin{figure}[h!]
\centering
\includegraphics[width=0.45\textwidth]{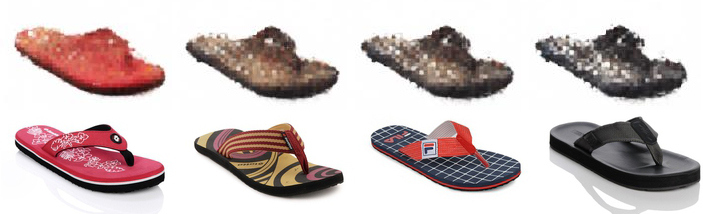}
\caption{Changing Attribute of Fashion Product Image by moving the sixth slider and its Corresponding Most Similar Item}
\label{fig:movingslider}
\end{figure}

Moving one slider changes some attribute of the the product image. In Figure \ref{fig:movingslider}, it illustrates how the product reconstructed image changes as the sixth slider is moved from left to right changing mostly its color and texture. 

In traditional Variational Autoencoders, it is difficult to determine which specific attributes of the image will transform upon changing one latent code in the $z$-vector. 

\subsection{Clustering the Dataset}
Each image from the training dataset is encoded using the encoder of the trained model and stored in the database. Both the $\mu_x$ and $\sigma_x$ vectors and fixed $z_x$ vectors are part of the encodings upon finishing training the model. These encodings will be used for retrieving similar images, clustering and recommendations.

K-means clustering with $k$ = 3 is used to cluster the entries in the database using both $\mu_x$ vector and fixed $z_x$ vector. Each entry has a subcategory tag (e.g. footwear, eyewear, or bag) and will be used for evaluating both clustering and similar image retrieval. All fashion products in the database are included in this clustering procedure. 

\begin{table}[h!]
\centering
\begin{tabular}{l c c c c}
\hline
Class  & Accuracy & Precision & Recall & F1 \\
\hline
Bags            & 0.863  & 1.0   &  0.69    & 0.81 \\
Footwear        & 0.985  & 0.98  &  0.98    & 0.98 \\
Eyewear         & 0.864  & 0.53  &  1.0     & 0.70 \\
\hline
\end{tabular}
\caption{Evaluation Metrics for Clustering using Fixed-Epsilon Sampled Encodings}
\label{tab:clusterfixed}
\end{table}

\begin{table}[h!]
\centering
\begin{tabular}{l c c c c}
\hline
Class  & Accuracy & Precision & Recall & F1 \\
\hline
Bags            & 0.866  & 1.0   &  0.69    & 0.82 \\
Footwear        & 0.985  & 0.98  &  0.98    & 0.98 \\
Eyewear         & 0.865  & 0.54  &  1.0     & 0.70 \\
\hline
\end{tabular}
\caption{Evaluation Metrics for Clustering using $\mu$-vector Encodings}
\label{tab:clustermu}
\end{table}

As seen in Table \ref{tab:clusterfixed} and \ref{tab:clustermu}, the F1 score of the two cases are almost similar. The F1 scores are decent considering that only a feature consisting of 16 dimensions is used to cluster images which resembles image classification although the scores are relatively lower on eyewear as K-means often tries to produce clusters of relatively uniform sizes and is very sensitive to skewed or imbalanced dataset \cite{imbalancedkmeans}. This means that using K-Means on this imbalanced dataset caused very low precision for eyewears and low recall for bags. This leads to having wrongly categorized items and cross-product recommendations that is not a member of the category. Even though each images have correct tags, the cluster assignments predicted by K-Means and the cluster centers shall be used.

\subsection{Getting Most Similar Images}
The most similar images are retrieved using log-likelihood maximization and Euclidean distance on fixed-epsilon sampling encoding as discussed earlier. But before retrieving similar images to the one generated by the user, its cluster is predicted first using Euclidean distance between $\bar{z}$-vector and the cluster centers. The application retrieves similar images from the cluster it was assigned.

\begin{table}[h!]
\centering
\begin{tabular}{l c c c c}
\hline
Config & Top-10 & Top-25 & Top-50 & Top-500 \\
\hline
Full / Log-Like                      & 0.946             & 0.918            &  0.894            & 0.762 \\
Full / Fixed-$\epsilon$              & \textbf{0.955}    & 0.947            &  0.899            & 0.846 \\
Cluster /Log-Like                    & 0.954             & 0.942            &  0.942            & 0.896 \\
Cluster / Fixed-$\epsilon$           & 0.953             & \textbf{0.950}   &  \textbf{0.948}   & \textbf{0.922} \\
\hline
\end{tabular}
\caption{mAP for Similar Image Retrieval on 1500 samples. \textbf{Full:} image retrieval on whole db; \textbf{Cluster:} retrieval on same cluster. Clustering is done with K-Means using fixed epsilon $z$-encodings (see \ref{tab:clustermu}}
\label{tab:map}
\end{table}

One evaluation metric for image retrieval is Mean Average Precision (mAP) \cite{designingvae}. Here, it is calculated on a randomly sampled 500 images for each category in retrieving top-10, top-25, top-50, and top-500 similar images. Figure \ref{tab:map} shows that using Euclidean Distance on Fixed-Epsilon Sampling Encodings yield higher mAP score most especially if the image retrieval is done on those which belong to the same cluster. Not only it is quicker but it has less false positives. This is highly encouraging since \cite{designingvae} had mAP score of 0.95 on their best model using VAE for MNIST images which are a less complex dataset.

\begin{figure}[h!]
\centering
\includegraphics[width=0.40\textwidth]{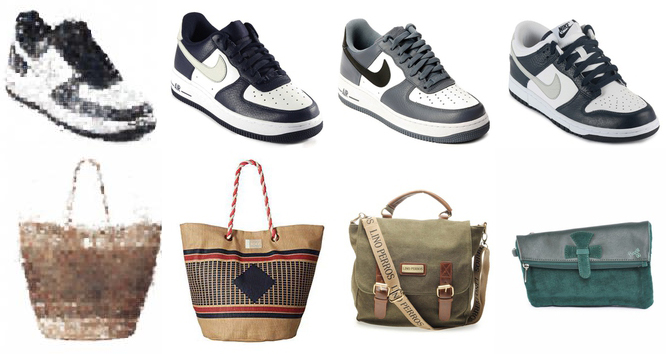}
\caption{Retrieved Most Similar Images using Log-Likelihood Maximization}
\label{fig:notfixedsim}
\end{figure}

\begin{figure}[h!]
\centering
\includegraphics[width=0.40\textwidth]{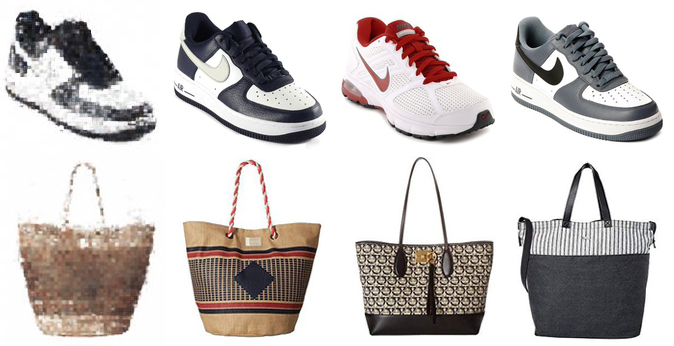}
\caption{Retrieved Most Similar Images using Euclidean Distance on Fixed-Epsilon Sampling Encodings}
\label{fig:fixedsim}
\end{figure}

Figure \ref{fig:notfixedsim} and \ref{fig:fixedsim} shows comparison of using these two methods for similar image retrieval. 
By observation, both techniques present the same image as the best match from the database but deviates on the second and third best match. From these two figures (Fig. \ref{fig:notfixedsim} and \ref{fig:fixedsim}), the first technique seems to prioritize similarities in color and texture while the second method tries to give emphasis on overall structure. 

\subsection{Recommending Products From Other Categories}
In recommending products from other categories, after getting \textit{$recommendation_{ij}$} (as discussed in  3.4), the closest image match from the database in that cluster is provided to the user using the two image matching techniques discussed in the paper so far. 

\begin{figure}[h!]
\centering
\includegraphics[width=0.45\textwidth]{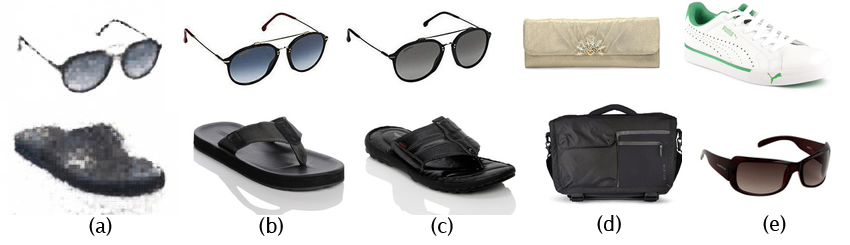}
\caption{Cross-Product Recommendation using Log-Likelihood Maximization for Similar Image Retrieval. (a) Generated Image, (b, c) Most Similar Match in the Cluster, (d, e) Recommendations from Other Clusters}
\label{fig:recommend_not_fixed}
\end{figure}

\begin{figure}[h!]
\centering
\includegraphics[width=0.45\textwidth]{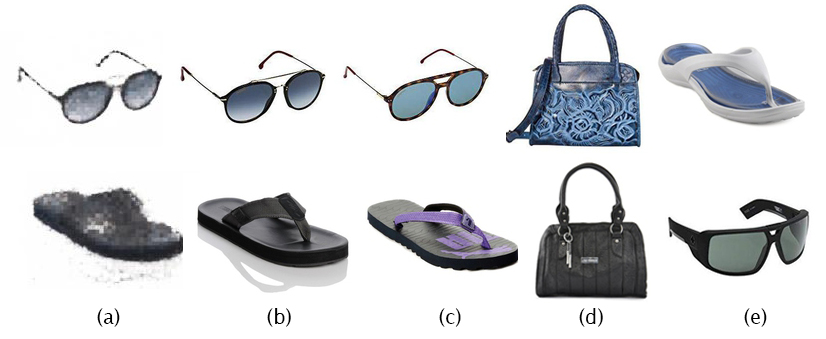}
\caption{Cross-Product Recommendation using Euclidean Distance on Fixed-Epsilon Sampling Encodings for Similar Image Retrieval. (a) Generated Image, (b, c) Most Similar Match in the Cluster, (d, e) Recommendations from Other Clusters}
\label{fig:recommend_fixed}
\end{figure}

Figures \ref{fig:recommend_not_fixed} and \ref{fig:recommend_fixed} illustrate sample recommendations for other clusters. Similar to the observation in Section 4.5, both similar image retrieval techniques for the cluster assignment of the generated product obtain the same output for the best match (b) but are different in the second best match (c). Recommendations using both methods are quite different from each other as observed both figures. 

Figure \ref{fig:recommend_fixed} shows the fixed-epsilon sampling technique seems to provide a better suggestion of products to pair with the eyeglasses emphasizing color and patterns recommending a blue bag and blue footwear to go with an eyewear with bluish tint. This is not the case recommendations for eyewear in Figure \ref{fig:recommend_not_fixed} which tries to suggest a bag and a shoe of mostly white in color. 

Both suggestions provide almost a similar items - a brown, leather bag and a thick-framed eyeglases. Fixed-epsilon sampling technique (Fig. \ref{fig:recommend_fixed} tries to recommend an eyeglass grayish tint similar to the generated image while the other method brings out the brown hue similar to the suggested bag and second best matching image.

\section{Conclusion}
In this work, the author starts with the idea that applications of deep learning in e-commerce, most especially in the fashion industry have been given more attention with the availability of new datasets. Product design generation, image retrieval mechanisms, and recommender systems are applications of deep learning but solutions only focus on using one architecture to solve one specific problem.

The author used Variational Autoencoders (VAEs) in a simple interactive application that aims to solve these problems with only one central architecture. Using the capability of VAEs in generating latent codes for generative modelling and feature extraction, new methods have been introduced on how to use these in generating product designs according to the user's preference, retrieve and cluster similar images, and recommend products from other categories. The user is able to design fashion products by their preference by adjusting a set of interactive sliders. Log-likelihood maximization and Euclidean Distance for Fixed Epsilon Sampling Encodings in retrieving similar and clustering using the latent codes showed promising results. Clustering with K-Means using these two methods showed good F1 score as high as 0.98 even with imbalanced data. Similar image retrieval using the latent codes also had astonishing results with mAP as high as 0.95 for top-10, top-25, and top-50 most similar image retrieval consistently scoring higher when using Fixed Epsilon Sampling Encodings compared to Log-Likelihood Maximization. The application is also able to use the user-generated latent codes, image retrieval techniques, and cluster centers from prior clustering for multiple cross-product recommendation. It was able to recommend items that are of the same attribute in color, texture, patterns, etc. with the product designed by the user. Overall, the pipeline was able to make full use of one deep learning architecture for multiple areas in e-commerce industry. 

\section*{Acknowledgments}
The author would like to express his gratitude to his professor, Rowel Atienza. for imparting him knowledge in this study.
The author would also like to express his gratitude and thanks to his colleagues and friends, especially Joanne Madridejos, who helped him in finishing the paper and the application.

\appendix

\bibliographystyle{named}
\bibliography{ijcai20}

\begin{thebibliography}{}

\bibitem[\protect\citeauthoryear{Chen \bgroup \em et al.\egroup
  }{2017}]{cnnrecom}
Luyang Chen, Fan Yang, and Heqing Yang.
\newblock Image-based product recommendation system with convolutional neural
  networks, 2017.

\bibitem[\protect\citeauthoryear{Deverall \bgroup \em et al.\egroup
  }{2017}]{shoegan}
Jaime Deverall, Jiwoo Lee, and Miguel Ayala.
\newblock Using generative adversarial networks to design shoes: the
  preliminary steps.
\newblock {\em CS231n in Stanford}, 2017.

\bibitem[\protect\citeauthoryear{Khan and Ahmad}{2004}]{kmeans}
Shehroz~S Khan and Amir Ahmad.
\newblock Cluster center initialization algorithm for k-means clustering.
\newblock {\em Pattern recognition letters}, 25(11):1293--1302, 2004.

\bibitem[\protect\citeauthoryear{Kingma and Welling}{2013}]{autoencoding}
Diederik~P Kingma and Max Welling.
\newblock Auto-encoding variational bayes.
\newblock {\em arXiv preprint arXiv:1312.6114}, 2013.

\bibitem[\protect\citeauthoryear{Krizhevsky and
  Hinton}{2011}]{autoencoderretrieval}
Alex Krizhevsky and Geoffrey~E Hinton.
\newblock Using very deep autoencoders for content-based image retrieval.
\newblock In {\em ESANN}, volume~1, page~2, 2011.

\bibitem[\protect\citeauthoryear{Kumar and Gupta}{2019}]{cgan}
Sudhir Kumar and Mithun~Das Gupta.
\newblock $ c^+ $ gan: Complementary fashion item recommendation.
\newblock {\em arXiv preprint arXiv:1906.05596}, 2019.

\bibitem[\protect\citeauthoryear{Kumar \bgroup \em et al.\egroup
  }{2014}]{imbalancedkmeans}
Ch~N~Santhosh Kumar, K~Nageswara Rao, A~Govardhan, and K~Sudheer Reddy.
\newblock Imbalanced k-means: An algorithm to cluster imbalanced-distributed
  data.
\newblock {\em International Journal of Engineering and Technical Research
  (IJETR)}, 2(2), 2014.

\bibitem[\protect\citeauthoryear{Lin \bgroup \em et al.\egroup
  }{2019}]{farmoutfit}
Yujie Lin, Pengjie Ren, Zhumin Chen, Zhaochun Ren, Jun Ma, and Maarten
  de~Rijke.
\newblock Improving outfit recommendation with co-supervision of fashion
  generation.
\newblock In {\em The World Wide Web Conference}, pages 1095--1105. ACM, 2019.

\bibitem[\protect\citeauthoryear{Liu \bgroup \em et al.\egroup
  }{2016}]{deepfashion}
Ziwei Liu, Ping Luo, Shi Qiu, Xiaogang Wang, and Xiaoou Tang.
\newblock Deepfashion: Powering robust clothes recognition and retrieval with
  rich annotations.
\newblock In {\em Proceedings of the IEEE conference on computer vision and
  pattern recognition}, pages 1096--1104, 2016.

\bibitem[\protect\citeauthoryear{Odaibo}{2019}]{tutorialderive}
Stephen Odaibo.
\newblock Tutorial: Deriving the standard variational autoencoder (vae) loss
  function.
\newblock {\em arXiv preprint arXiv:1907.08956}, 2019.

\bibitem[\protect\citeauthoryear{Qayyum \bgroup \em et al.\egroup
  }{2017}]{medicalimage}
Adnan Qayyum, Syed~Muhammad Anwar, Muhammad Awais, and Muhammad Majid.
\newblock Medical image retrieval using deep convolutional neural network.
\newblock {\em Neurocomputing}, 266:8--20, 2017.

\bibitem[\protect\citeauthoryear{ReNom}{}]{ReNom}
ReNom.
\newblock Variational auto-encoder.

\bibitem[\protect\citeauthoryear{Rostamzadeh \bgroup \em et al.\egroup
  }{2018}]{fashiongen}
Negar Rostamzadeh, Seyedarian Hosseini, Thomas Boquet, Wojciech Stokowiec, Ying
  Zhang, Christian Jauvin, and Chris Pal.
\newblock Fashion-gen: The generative fashion dataset and challenge.
\newblock {\em arXiv preprint arXiv:1806.08317}, 2018.

\bibitem[\protect\citeauthoryear{Spinner \bgroup \em et al.\egroup
  }{2018}]{towardsvae}
Thilo Spinner, Jonas K{\"o}rner, Jochen G{\"o}rtler, and Oliver Deussen.
\newblock Towards an interpretable latent space: an intuitive comparison of
  autoencoders with variational autoencoders.
\newblock In {\em IEEE VIS 2018}, 2018.

\bibitem[\protect\citeauthoryear{Torres~Fernandez}{2018}]{designingvae}
Sara Torres~Fernandez.
\newblock Designing variational autoencoders for image retrieval, 2018.

\bibitem[\protect\citeauthoryear{Tuinhof \bgroup \em et al.\egroup
  }{2018}]{cnndeepreco}
Hessel Tuinhof, Clemens Pirker, and Markus Haltmeier.
\newblock Image-based fashion product recommendation with deep learning.
\newblock In {\em International Conference on Machine Learning, Optimization,
  and Data Science}, pages 472--481. Springer, 2018.

\bibitem[\protect\citeauthoryear{Zhu \bgroup \em et al.\egroup
  }{2018}]{vaegeneration}
Jiating Zhu, Yu~Yang, Jiannong Cao, and Esther Chak~Fung Mei.
\newblock New product design with popular fashion style discovery using machine
  learning.
\newblock In {\em International Conference on Artificial Intelligence on
  Textile and Apparel}, pages 121--128. Springer, 2018.

\bibitem[\protect\citeauthoryear{Zou \bgroup \em et al.\egroup
  }{2019}]{fashionai}
Xingxing Zou, Xiangheng Kong, Waikeung Wong, Congde Wang, Yuguang Liu, and Yang
  Cao.
\newblock Fashionai: A hierarchical dataset for fashion understanding.
\newblock In {\em Proceedings of the IEEE Conference on Computer Vision and
  Pattern Recognition Workshops}, pages 0--0, 2019.

\end{thebibliography}

\end{document}